\documentclass{article}
\usepackage{graphicx}     
\usepackage{float}        
\usepackage{url}
\usepackage{hyperref}

\providecommand{\keywords}[1]{\textbf{\textit{Keywords---}} #1}

\usepackage{subcaption}   
\usepackage{multirow}
\usepackage[utf8]{inputenc}
\usepackage{float} 

\title{Real-Time Fitness Exercise Classification and Counting from Video Frames}
\author{
  Riccardo Riccio \\
  Department of Mathematics, University of Pavia, Pavia, Italy \\
  \texttt{riccardo.riccio01@universitadipavia.it}
}

\date{}

\begin{document}

\maketitle

\begin{abstract}
This paper introduces a novel method for real-time exercise classification using a Bidirectional Long Short-Term Memory (BiLSTM) neural network. Existing exercise recognition approaches often rely on synthetic datasets, raw coordinate inputs sensitive to user and camera variations, and fail to fully exploit the temporal dependencies in exercise movements. These issues limit their generalizability and robustness in real-world conditions, where lighting, camera angles, and user body types vary.

To address these challenges, we propose a BiLSTM-based model that leverages invariant features—joint angles in addition to raw coordinates. By using both angles and (x, y, z) coordinates, the model adapts to changes in perspective, user positioning, and body differences, improving generalization. Training on 30-frame sequences enables the BiLSTM to capture the temporal context of exercises and recognize patterns evolving over time.

We compiled a comprehensive dataset by combining synthetic data from the InfiniteRep dataset and real-world videos from the Kaggle Workout/Exercises Video Dataset and other online sources. This dataset includes four common exercises: squat, push-up, shoulder press, and bicep curl. The model was trained and validated on these diverse datasets, achieving an accuracy of over 99\% on the test set. To assess generalizability, the model was tested on 2 separate test sets representative of typical usage conditions. Comparisons with the previous approach from the literature are present in the result section showing that the proposed model is the best-performing one.

Furthermore, the classifier was integrated into a user-friendly web application that provides real-time exercise classification and repetition counting without the need for manual exercise selection.
Demo and datasets available at: \href{https://github.com/RiccardoRiccio/Fitness-AI-Trainer-With-Automatic-Exercise-Recognition-and-Counting}{GitHub Repository}.

\end{abstract}


\section{Introduction}

Exercise is a critical component of a healthy lifestyle, contributing to both physical fitness and mental well-being. Despite its well-documented benefits, many individuals struggle to maintain consistent workout routines. Common barriers include a lack of knowledge about proper exercise techniques and challenges in tracking workout progress. Traditional methods, such as manual repetition counting or feedback from personal trainers, are often inaccessible or inconvenient for many users. These difficulties can lead to poor adherence to exercise routines, ultimately affecting overall health outcomes.

Recent advancements in artificial intelligence (AI) and computer vision offer promising solutions to these challenges. In particular, pose estimation techniques, which identify and track key body landmarks from images or videos, have emerged as powerful tools for analyzing human movement. By combining pose estimation with machine learning (ML) models, it is possible to develop systems that automatically count repetitions, correct form, and even classify the exercise being performed. Such systems have the potential to revolutionize fitness tracking, making exercise more accessible, effective, and engaging for users.

However, existing approaches to exercise classification and repetition counting often suffer from significant limitations. Many models, such as those relying on convolutional neural networks (CNNs) with voting mechanisms or simple frame-based classifiers, treat individual frames independently, failing to fully capture the temporal dynamics of exercise movements. This leads to reduced classification accuracy, particularly for exercises that share similar initial postures. Furthermore, methods that utilize raw (x, y, z) coordinates of joints are often sensitive to variations in user positioning, camera angles, and distances, limiting their generalizability to real-world environments.

To address these limitations, this paper proposes a novel approach to automatic exercise classification using a bidirectional Long Short-Term Memory (BiLSTM) architecture. The BiLSTM model is designed to capture the sequential nature of exercise movements by processing frames in both forward and backward directions, allowing it to better distinguish between exercises that may appear similar at individual points in time. Unlike models that rely solely on raw coordinates, the proposed approach leverages angles between key joints, which are invariant to changes in camera perspective and user positioning, thus improving robustness and generalization.

While state-of-the-art models like Vision Transformers [1] and more recent pose estimation methods exist, this project aims to develop a model capable of real-time classification and counting of exercises in resource-constrained environments, such as mobile devices. Thus,  computational efficiency was prioritized while maintaining high accuracy. The BiLSTM and BlazePose models were chosen to balance these trade-offs, as they offer excellent performance with significantly reduced computational demands, which is essential for real-time applications.

The core contribution of this paper is the development of an automatic exercise classification system capable of identifying four common exercises without requiring manual input from the user and improving current methods to leverage invariant features and the sequentiality of the data. The model was trained on a combination of real-world and synthetic datasets, ensuring robustness across diverse conditions. Additionally, multiple test sets were created to assess the model's generalization capabilities, demonstrating its effectiveness in both controlled and real-world scenarios. The system is designed to perform in diverse environments, from home workouts to gym settings, and is integrated into a web application, allowing for real-time exercise classification and repetition counting via a webcam, showcasing its practical applicability.

In this paper, we evaluate the performance of the BiLSTM model on both synthetic and real-world data, demonstrating its effectiveness in handling the temporal dynamics of exercise movements and its robustness in real-world conditions. 

\section{Related Works}

In the domain of exercise classification, leveraging pose estimation and machine learning has become a pivotal approach to identifying physical movements in real-time. This section reviews the relevant literature on pose estimation and automatic exercise classification, with a focus on methodologies, model architectures, and how this work improves upon existing limitations.

\subsection{Human Pose Estimation for Exercise Tracking}

Pose estimation is a fundamental aspect of understanding human movement. It involves detecting key body landmarks such as joints and limbs to represent a person's posture in either two or three dimensions. This technology has found applications across fields like fitness, healthcare, and sports. One of the more recent advancements in pose estimation is the use of deep learning models, which have significantly improved accuracy and robustness.

For this project, Mediapipe's BlazePose [2, 3] was selected as the primary pose estimation framework. BlazePose is specifically optimized for real-time performance on mobile devices and was chosen for its ability to efficiently track body landmarks. BlazePose predicts 33 key body points, including the head, shoulders, elbows, wrists, hips, knees, and ankles, making it a suitable choice for fitness applications. The model also incorporates a detector-tracker pipeline to maintain accurate tracking of body parts throughout the sequence of movements.

The model needs to be developed with practical usage in mind, and while we have implemented it within a web application, a more natural future use case could be in a mobile app acting as an AI personal trainer. Given the need for low computational requirements to ensure real-time performance on mobile devices, BlazePose is an ideal choice, as it was created specifically with real-time mobile usage in mind. This makes it highly suited for applications where users require fast and accurate feedback during their workouts, even on devices with limited processing power.

\begin{figure}[H]
    \centering
    \includegraphics[width=0.7\textwidth, height=7cm]{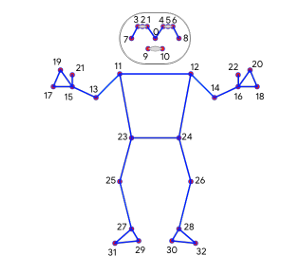} 
    \caption{topology of the 33 key points [2]}
    \label{fig:blazepose}
\end{figure}

\subsection{Automatic Exercise Classification}

Automatic exercise classification in real-time involves the use of machine learning models to analyze data extracted from video frames or sensor inputs to identify the exercise a user is performing. This paper focuses on using features that can be extracted from frames with a simple webcam, eliminating the need for additional sensors and relying solely on visual data. By utilizing pose estimation techniques to capture joint angles and relative distances directly from standard webcam input, the model achieves real-time exercise classification based on these visual features alone. Given the sequential nature of most physical exercises, effective classification typically requires models that can capture and interpret temporal dependencies in the data. This section reviews several methodologies from the literature that address this challenge, focusing on three papers [4, 5, 6]. Following this review, the limitations of these models are discussed, and how this thesis aims to address these challenges.

\subsubsection{Different Approaches to Automatic Exercise Classification}

\paragraph{1. CNN + Soft Voting:}
The paper [4] analyzed various approaches to handle exercise classification utilizing a synthetic dataset from Infinity AI [7], which includes videos of avatars performing 10 different exercises. Initially, the study explored traditional machine learning models like K-Nearest Neighbors (KNN) and Random Forest, but they showed limited performance due to their inability to capture temporal dependencies.

To better handle the temporal aspects of exercise movements, the study experimented with an LSTM model, which is suited for sequence prediction tasks. The LSTM model was trained on sequences of frames, incorporating standardized joint coordinate data and relative joint positions. This approach allowed the model to consider how movements evolved over time, improving classification accuracy with respect to traditional ML algorithms.

Building on this, the study introduced a hybrid model that combined CNNs and LSTMs. The CNN component was used to capture spatial features from each frame, while the LSTM processed these features over a sequence of frames to learn the temporal dynamics. This combined model slightly enhanced classification performance.

Finally, the study proposed an ensemble learning approach, where predictions from multiple CNN models, each analyzing a sequence of 30 frames, were aggregated using a soft voting mechanism. This ensemble method aimed to reduce the impact of anomalies and outliers by combining the outputs of different models, and achieved a significant improvement in classification accuracy, reaching 92.12\% (compared to 79.78\% of the combined CNN and LSTM).

\paragraph{2. LSTM using x,y,z coordinates:}
The paper [5] presents a real-time exercise classification system that utilizes keypoint data extracted from video feeds. The system uses MediaPipe’s BlazePose to identify and track 33 key landmarks on the human body, capturing these as (x, y, z) coordinates along with confidence levels for each point. These key points are then used as input to a stacked Long Short-Term Memory (LSTM) neural network, which is particularly suited for processing sequential data.

The LSTM model is designed to capture the temporal dynamics of exercise movements by processing a sequence of frames. For this system, 8 consecutive frames are used to give the temporal context to classify the exercise being performed. This approach allows the model to recognize the flow of movement across time, which is crucial for distinguishing between exercises that may appear similar in individual frames but differ in their sequence of motions.

The model was trained using a subset of the UCF101 Human Actions dataset [8], which contains video clips of various activities. The authors selected four specific exercises (push-ups, lunges, bodyweight squats, and discus throws) for training and validation. The dataset provides a diverse range of movements, ensuring the model can learn to classify exercises with different body orientations and motion ranges.

\paragraph{3. DNN on single frame + aggregation:}
In the paper [6], the model developed is designed to classify 3 exercises (push-ups, pull-ups, and squats). The input to the model consists of 3D human body key points, which are extracted from each frame of the video using the OpenPose model [9]. These key points represent the skeletal structure of the person, capturing the positions of various joints in three dimensions.

The model’s architecture is built around an action recognition neural network that uses these key points to classify the exercise. For each video frame, the model generates a prediction about which exercise is being performed.

The model processes the video frame by frame, making an individual prediction for each one. After 10 frames, it aggregates these predictions using a sliding window and selects the most common exercise identified across these frames as the final classification for that segment.

\subsection{Limitations of These Approaches and the Model Proposed in This Project}

The CNN + Soft Voting approach, as detailed in the paper, relies heavily on a synthetic dataset (Infinity AI’s Fitness Basic Dataset). This reliance raises concerns about the model’s ability to generalize to real-world data, where variability in lighting, camera angles, and user body types is more prevalent. Although the final model combines CNNs with a soft voting mechanism across 30 frames, it doesn’t fully exploit the temporal dependencies inherent in exercise movements. The model analyzes each frame somewhat independently, and the final classification is based on an average of these predictions. This approach does not fully leverage the sequential nature of exercise movements.

The LSTM-based approach using (x, y, z) coordinates introduces challenges related to invariance. By using raw (x, y, z) coordinates as input, the model might struggle when used with data taken from a different distribution. For example, if the dataset contained exercises performed at a fixed distance from the camera, using the coordinates might not generalize well to exercises performed by users at different distances, on different screen dimensions, or with varying user heights. Angles are invariant to the user’s position relative to the camera, which can make the model more robust to variations in camera perspective and scale. Additionally, the use of only 8 consecutive frames to capture temporal dynamics may not provide sufficient context for exercises with complex or longer movement sequences, potentially reducing classification accuracy for more intricate activities.

In the Deep Learning approach, the problem is again that each frame is treated as an independent entity. This approach fails to capture the sequential nature of exercise data. By making predictions based on single frames and then aggregating these predictions by selecting the most common outcome across 10 frames, the model does not fully utilize the temporal continuity of the data. Also, this paper used the coordinates, which, as explained before, have the invariance problem.

To address the limitations, the BiLSTM model proposed in this thesis leverages the temporal nature of exercise data by analyzing sequences of frames both forward and backward. Unlike models that treat frames independently, the BiLSTM captures the complete temporal context, enabling it to recognize patterns that evolve over time. The proposed model utilizes additional angle features derived from key joint positions. Additionally, the model processes 30 frames at a time, balancing the need for sufficient sequential information with the requirements of real-time performance. The dataset used for this thesis includes both synthetic and real-world video data to improve the model’s generalization capabilities.

The BiLSTM model was chosen for its ability to effectively capture temporal dependencies while maintaining the computational efficiency needed for real-time applications, unlike more computationally intensive architectures. Additionally, the model achieved nearly perfect performance on the validation set, as a consequence there was no need to rely on additional models.

In real-time applications, especially those running on resource-limited hardware like mobile phones, the balance between computational efficiency and accuracy is critical. Similarly, BlazePose was selected over more complex pose estimation models due to its optimization for real-time performance on mobile devices, further aligning with the project’s objective.

\section{Methodology}

This section outlines the approach taken to develop the exercise classifier, from data collection, feature extraction, model selection, and training processes. The classifier is built using a combination of real and synthetic datasets, designed to reflect the diverse environments in order to enhance its generalizability. The model is intended for integration into a real-time fitness tracking application, where it will automatically recognize the exercises the user performs in front of a webcam.
The section begins with an overview of the datasets used for training, emphasizing the methods employed to capture realistic exercise scenarios. It then details the landmark extraction process, which utilizes pose estimation to extract body coordinates from exercise videos. Then, the feature extraction and preprocessing steps are discussed, explaining how raw landmark data is transformed into suitable inputs for models. The final part covers the selection and training of LSTM and Bidirectional LSTM models and evaluates their performance across various test conditions to ensure their effectiveness and reliability in practical usage.

\subsection{Dataset}

The creation of a suitable dataset requires data that tends to resemble the conditions under which the application will be used. Ideally, the data should include videos where the user’s head and main joints are clearly visible, as BlazePose utilizes face detection to initiate tracking of the person’s body. The exercises should also be performed in a manner similar to how they will be executed later in the app.  The creation of the dataset was not trivial, as while the user is advised to use the application in a controlled environment like home, it is likely that the usage will also extend to gyms or other places. Therefore, the dataset was designed to account for this variability, ensuring the model performs well in diverse environments.
In order to build the current dataset, three sources were combined:

The first source is the “Kaggle Workout/Exercises Video Dataset” [10], which includes videos of various gym exercises performed by expert trainers coming mainly from YouTube. It originally contained various videos of  22 different exercises but just 4 were used (barbell bicep curl, squat, push up and shoulder press). This dataset initially contained only 19 squat videos, while other exercises had around 50 videos each. To maintain balance, additional squat videos have been added taken from other online sources [11, 12, 13, 14] and for the other exercise only a subset has been used, resulting in 25 videos per exercise. This dataset provides a mix of real-world exercise scenarios with variations in backgrounds and lighting conditions.

The second source, the InfiniteRep Dataset [7], is composed of a synthetic video dataset where human-like avatars perform various exercises. For each class, 100 videos were selected, creating a balanced dataset that includes diverse variations in body shapes, lighting, and camera angles. This dataset has been selected since it resembles more the one the user will perform. Moreover, it enables to increase significantly the dimension of the dataset which is important for improving performance for the DL models, in particular since it's not easy to find ready-to-use datasets for these exercises.

\begin{figure}[H]
\centering
\begin{subfigure}[b]{0.3\textwidth}
    \centering
    \includegraphics[height=4cm, width=4.5cm]{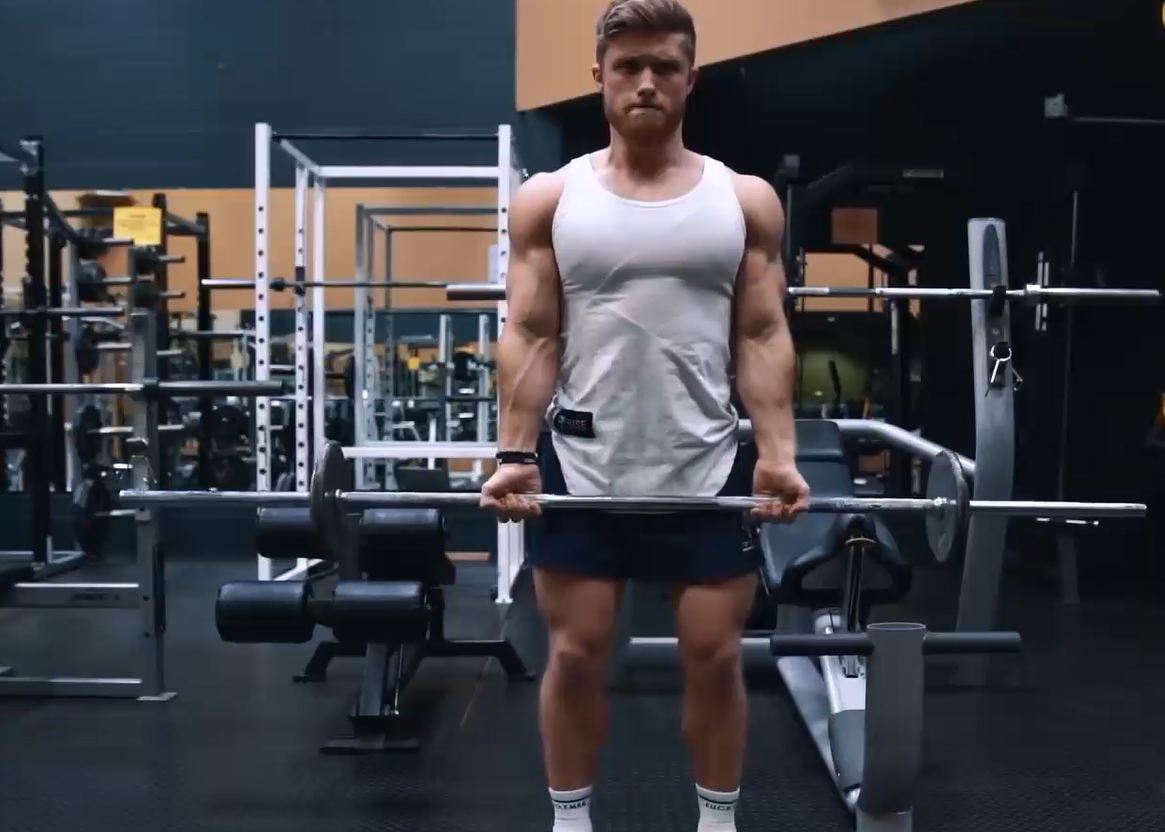} 
    \caption{Kaggle Dataset}
    \label{fig:kaggle_dataset}
\end{subfigure}
\hfill
\begin{subfigure}[b]{0.3\textwidth}
    \centering
    \includegraphics[height=4cm]{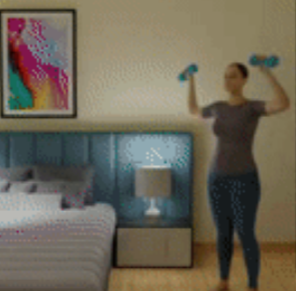} 
    \caption{InfiniteRep Dataset}
    \label{fig:infinite_rep}
\end{subfigure}
\hfill
\begin{subfigure}[b]{0.3\textwidth}
    \centering
    \includegraphics[height=4cm]{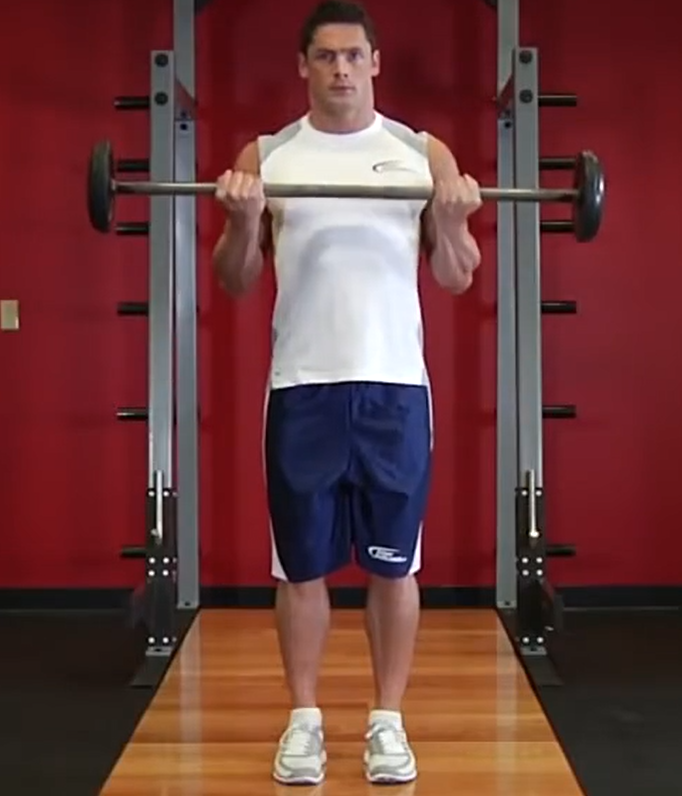} 
    \caption{Similar Dataset}
    \label{fig:online_sources}
\end{subfigure}
\caption{Images from the Datasets used for training and validation of the exercise classifier. (images are cropped due to screenshots, hence they appear to have different dimensions)}
\label{fig:datasets}
\end{figure}

Lastly, additional videos were compiled from various free online sources (as before) to further align the data with the expected use of the app. These videos feature exercises that more closely mirror the intended conditions of the app but still contain significant variability in angle and environment, simulating realistic and diverse exercise conditions.

In order to evaluate better the performance of the models, two other datasets have been created just for testing:

The "Final My Test Video" dataset includes 4 videos per exercise class, recorded at home to closely resemble the typical user experience. These videos are representative of how users would use the app, with clear visibility of the body and face, and primarily frontal or slightly angled views. The "Final Test Gym Video" dataset consists of 5 videos per exercise class, recorded in a gym or similar setting. This second test set contains a video with more difficult angulations and occlusions, to test better the model generalization capabilities. Although the app is primarily recommended for home use, this dataset evaluates the model’s ability to handle more variable conditions.

\begin{figure}[H]
\centering
\begin{subfigure}[b]{0.4\textwidth} 
    \centering
    \includegraphics[height=4cm]{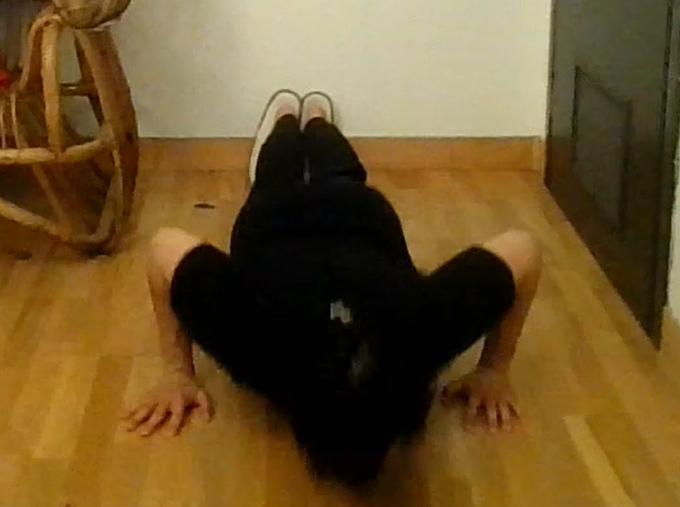} 
    \caption{Final My Test Video}
    \label{fig:mytestvideo}
\end{subfigure}
\hspace{1cm} 
\begin{subfigure}[b]{0.4\textwidth} 
    \centering
    \includegraphics[height=4cm]{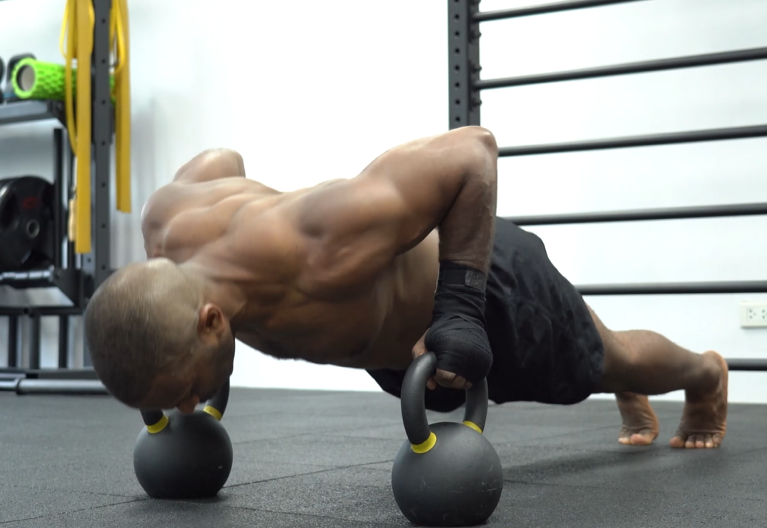} 
    \caption{Final Test Gym Video }
    \label{fig:tesygym}
\end{subfigure}
\caption{Images of the datasets used for additional testing (images are cropped due to screenshots)}
\label{fig:additional_images}
\end{figure}

By combining these datasets, the exercise classifier was trained and tested on data that accurately represents the diverse conditions under which the app will be used. This approach enhances the model's ability to generalize and perform accurately across various real-world environments, ensuring that it can assist users regardless of their workout location.

\subsection{Preprocessing Pipeline for Feature Creation}

The following section describes the preprocessing pipeline used to create the features for training, outlining how information is extracted from video frames to then make predictions. The choice of features was based on some important considerations:

The primary consideration in feature selection was to ensure that the model receives sufficient information to differentiate between the various exercises while also keeping the feature set manageable for offline use. It is essential to limit the number of features because LSTM models can struggle with sequences that are too long. Moreover, the features should be invariant or have minimal variance concerning the different conditions under which the application might be used. For instance, users may utilize webcams at varying distances or have different physical characteristics such as height and weight.

The key decision was to avoid relying on only the plain coordinates but use a mixture of both plain coordinates and angles. The assumption is that the angle gives the invariance to a new scenario while adding also the coordinate might help in cases with more occlusion and missing landmarks. On top of that using features similar to [4] make the comparison easier and more meaningful with the BiLSTM proposed.
\subsubsection{Landmark and Features Extraction}

The landmark extraction process begins by organizing the datasets into folders, each containing subfolders for the four exercises: barbell bicep curl, shoulder press, squat, and push-up. The extraction process iterates over each video and processes each frame using pose estimation with MediaPipe to detect landmarks.

For each frame, the process extracts (x, y, z) coordinates of a subset of relevant landmarks, including shoulders, elbows, wrists, hips, knees, and ankles. The extraction process ensures that essential landmarks specific to each exercise are detected on at least one side (left or right). If critical landmarks are missing on both sides (indicated by checks on left-side-valid and right-side-valid functions), the frame is skipped to ensure that only complete data is used for later processing. In cases where some essential landmarks are not detected, placeholder values [0.0, 0.0, 0.0] are used to maintain a consistent data structure.

When a frame passes these checks, 22 landmarks, along with the video ID and exercise type (label), are recorded and later stored in a CSV file named according to the dataset. From this landmark 12 angles are extracted.
At this stage, the CSV file will contain a number of rows equal to the valid frames and as columns the video ID, the exercise label, and (x, y, z) coordinates for each of the 22 relevant landmarks (so, the CSV file will have a shape of (number of valid frames, 80). This structured approach helps organize data for then being preprocessed in order to be ready for the model’s training (see Tables 1 and 2 for a list of all the features).

 This strategy helps maintain the dataset's structure and ensures that missing data does not disrupt the model's learning process.

\begin{table}[H]
    \centering
    \begin{minipage}[t]{0.48\textwidth} 
        \centering
        \caption{Angles}
        \renewcommand{\arraystretch}{1} 
        \setlength{\tabcolsep}{1pt} 
        \small 
        \begin{tabular}{|p{0.99\linewidth}|} 
            \hline
            \textbf{Landmarks} \\ \hline
            LEFT\_HIP, LEFT\_SHOULDER, LEFT\_ELBOW \\ \hline
            RIGHT\_HIP, RIGHT\_SHOULDER, RIGHT\_ELBOW \\ \hline
            LEFT\_SHOULDER, LEFT\_ELBOW, LEFT\_WRIST \\ \hline
            RIGHT\_SHOULDER, RIGHT\_ELBOW, RIGHT\_WRIST \\ \hline
            LEFT\_HIP, LEFT\_KNEE, LEFT\_ANKLE \\ \hline
            RIGHT\_HIP, RIGHT\_KNEE, RIGHT\_ANKLE \\ \hline
            LEFT\_SHOULDER, LEFT\_HIP, LEFT\_KNEE \\ \hline
            RIGHT\_SHOULDER, RIGHT\_HIP, RIGHT\_KNEE \\ \hline
            LEFT\_KNEE, LEFT\_ANKLE, LEFT\_HEEL \\ \hline
            RIGHT\_KNEE, RIGHT\_ANKLE, RIGHT\_HEEL \\ \hline
            LEFT\_ANKLE, LEFT\_HEEL, LEFT\_FOOT\_INDEX \\ \hline
            RIGHT\_ANKLE, RIGHT\_HEEL, RIGHT\_FOOT\_INDEX \\ \hline
        \end{tabular}
        \label{tab:angles}
    \end{minipage}%
    \hfill 
    \begin{minipage}[t]{0.48\textwidth} 
        \centering
        \caption{Coordinates}
        \renewcommand{\arraystretch}{1} 
        \setlength{\tabcolsep}{2pt} 
        \small 
        \begin{tabular}{|p{0.9\linewidth}|} 
            \hline
            \textbf{Landmarks} \\ \hline
            LEFT\_SHOULDER, RIGHT\_SHOULDER \\ \hline
            LEFT\_HIP, RIGHT\_HIP \\ \hline
            LEFT\_KNEE, RIGHT\_KNEE \\ \hline
            LEFT\_ELBOW, RIGHT\_ELBOW \\ \hline
            LEFT\_WRIST, RIGHT\_WRIST \\ \hline
            LEFT\_ANKLE, RIGHT\_ANKLE \\ \hline
            LEFT\_HEEL, RIGHT\_KNEE \\ \hline
            LEFT\_HOOT, LEFT\_FOOT \\ \hline
            LEFT\_PINKY, RIGHT\_PINKY \\ \hline
            LEFT\_INDEX, RIGHT\_INDEX \\ \hline
            RIGHT\_THUMB, RIGHT\_THUMB \\ \hline
        \end{tabular}
        \label{tab:coordinates}
    \end{minipage}
\end{table}

To prepare the data for training, the features extracted from frames are aggregated into windows of 30 consecutive frames from the same video. This aggregation assumes that each video contains only one type of exercise, simplifying the task of classifying the exercise type. This is ensured by selecting a dataset that contains videos with only one exercise being performed. By grouping 30 frames together, the dataset captures a sequence of movement that reflects the exercise's dynamics. Each window is then labeled with the exercise type of the first frame in the window.
This structure ensures that the data is temporally consistent, enabling the model to learn from sequences of frames that represent continuous movement from the same exercise.

\subsection{Models Training and Evaluations}

The training phase involves using LSTM and BiLSTM to develop the exercise classification model. These models are well-suited for sequence prediction tasks, making them ideal for understanding the temporal dynamics of exercise movements from the aggregated landmark features. The training process begins with loading and combining preprocessed datasets, followed by hyperparameter tuning and evaluation to select the best-performing model with the optimal hyperparameters.

\subsubsection{Loading and Preparing Data}

First, the preprocessed features and labels from multiple datasets, including the “Kaggle Workout/Exercises Video Dataset”, “InfinitRep”, and “similar” datasets, are loaded and concatenated. This ensures that the final dataset is comprehensive and captures a wide range of variations in exercise performance, such as differences in body size, environment, and angles.

To prepare the data for the LSTM and BiLSTM models, labels are encoded using a label encoder, converting categorical exercise types into numerical values. These encoded labels are further transformed into a categorical format suitable for multi-class classification. Feature scaling is also performed using a standard scaler to normalize the data, which helps improve the convergence speed and overall training.

The dataset is then reshaped to match the input required by the LSTM. In particular, the data is reshaped into three dimensions: samples, timesteps, and features, resulting in a shape of (\textit{number\_samples}, 30, 78). 

\subsubsection{Model Architecture and Training}

Two models were developed for the exercise classification task: a standard LSTM model and a Bidirectional LSTM model. The LSTM model uses two LSTM layers with dropout layers in between to reduce overfitting. 
The BiLSTM model enhances the LSTM architecture by processing the sequence data in both forward and backward directions, allowing the model to capture patterns that may depend on future as well as past context. 

\begin{figure}[ht]
\centering
\begin{subfigure}[b]{0.4\textwidth} 
    \centering
    \includegraphics[height=7cm]{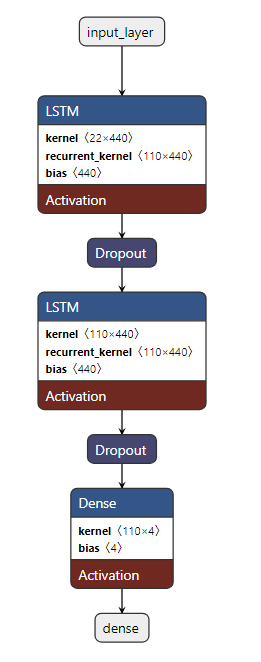} 
    \caption{LSTM model architecture (using netron.app [15])}
    \label{fig:lstmmodel}
\end{subfigure}
\hspace{1cm} 
\begin{subfigure}[b]{0.4\textwidth} 
    \centering
    \includegraphics[height=7cm]{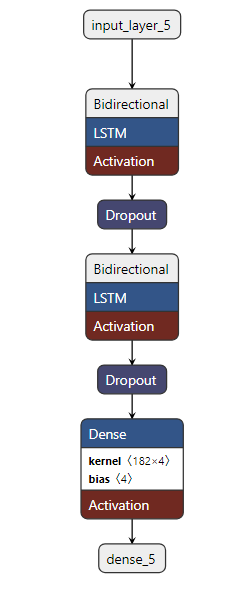} 
    \caption{BiLSTM model architecture (using netron.app)}
    \label{fig:bilstmmodel}
\end{subfigure}
\caption{model architecture}
\label{fig:modelarchitecture}
\end{figure}

\subsubsection{Hyperparameter Tuning}

Hyperparameter tuning was performed using a random search strategy (using 20 iteration), testing various combinations of key hyperparameters such as the number of LSTM units, dropout rate, learning rate, batch size, and the number of training epochs. Early stopping and learning rate reduction techniques were also employed during training. 

\begin{table}[H]
\centering
\caption{Hyperparameters Used for Both Models Tuning}
\begin{tabular}{|l|l|l|}
\hline
\textbf{Hyperparameter} & \textbf{Distribution}            & \textbf{Range}                  \\ \hline
\textbf{Units}          & Random Integer                   & 50 to 150                       \\ \hline
\textbf{Dropout Rate}   & Uniform Distribution             & 0.2 to 0.5                      \\ \hline
\textbf{Learning Rate}  & Uniform Distribution             & 0.0001 to 0.001                 \\ \hline
\textbf{Batch Size}     & Random Integer                   & 32 to 64                        \\ \hline
\textbf{Epochs}         & Random Integer                   & 50 to 100                       \\ \hline
\end{tabular}
\label{tab:hyperparameters}
\end{table}

\subsubsection{Model Evaluation}
During the training phase, the dataset is split into training, validation and, test sets (70-15-15) to assess the model's performance on unseen data.

To ensure that the models generalized well beyond the training data, both the LSTM and BiLSTM models were further tested on two additional datasets designed to reflect real-world conditions: the "Final My Test Video" dataset, which included exercises recorded at home, and the "Final Test Gym Video" dataset, which contained exercises recorded in gym or other environments. Testing on these datasets is useful since it simulates the various settings where the app might be used, such as homes, gyms, or other environments with different lighting and backgrounds.

The models were evaluated using the accuracy metric on the test set, and the best-performing model and hyperparameters were selected based on this metric. The primary evaluation metrics used were accuracy, precision, recall, and F1-score, which were detailed in classification reports for each model. These metrics provided a clear view of how well the models distinguished between different exercise classes. The evaluation also included confusion matrices, helping identify any specific exercises that the models might confuse with each other.

\section{Results}

This chapter presents a summary of the experiments and results conducted to evaluate the BiLSTM model proposed (with both coordinates and angles). It includes comparisons with an LSTM model utilizing the same feature set, a BiLSTM model that employs only coordinate data, and a BiLSTM model that leverages solely angles and normalized distances. Additionally, the chapter contrasts these models with the approaches discussed in the literature review, incorporating necessary modifications to facilitate a meaningful comparison.
The final automatic exercise classification model was used on the “Auto Classify” page of the web app (described in a later section). The evaluation of the repetition counting logic is not included, as it is based on a function that uses angles rather than AI. However, it might be beneficial to have the repetition counting logic reviewed by expert trainers or to expand the logic to cover additional perspectives, such as side angles, rather than just the frontal view with a slight angle. Similarly, the evaluation of the chatbot design is considered beyond the scope of this paper.

\subsection{Experimental Setup}

The experiments were conducted as follows: both LSTM and BiLSTM models were trained using the main dataset, with the best hyperparameters identified through tuning. The models with these optimized hyperparameters were then validated on the main dataset, and learning curves were plotted. Finally, both models were evaluated on two test datasets: “Final My Test Video” and “Final Test Gym Video.” The evaluation metrics used included accuracy, classification report, and confusion matrix.

\subsection{Best Hyperparameters}

 The table below shows the best hyperparameters found for each model. These models, configured with the optimized hyperparameters, were saved and used for the rest of the evaluations.

\begin{table}[H]
    \centering
    \caption{Best Hyperparameters for LSTM and BiLSTM Models}
    \label{tab:best-hyperparameters}
    \begin{tabular}{|l|c|c|}
        \hline
        \textbf{Hyperparameter} & \textbf{LSTM} & \textbf{BiLSTM} \\
        \hline
        Batch Size & 38 & 54 \\
        Dropout Rate & 0.3829 & 0.2174 \\
        Epochs & 57 & 73 \\
        Learning Rate & 0.0001 & 0.0004 \\
        Units & 117 & 73 \\
        \hline
    \end{tabular}
\end{table}

\subsection{Evaluation on Test Sets on The Dataset Used For Training}

The following tables present the accuracy, classification report, and confusion matrix for the test set.

\begin{table}[H]
    \centering
    \caption{Classification Report for LSTM Model}
    \label{tab:lstm-classification-report}
    \begin{tabular}{|l|c|c|c|c|}
        \hline
        \textbf{Class} & \textbf{Precision} & \textbf{Recall} & \textbf{F1-Score} & \textbf{Support} \\
        \hline
        Barbell Biceps Curl & 1.00 & 0.98 & 0.99 & 193 \\
        Push-up & 1.00 & 0.97 & 0.99 & 158 \\
        Squat & 0.97 & 0.99 & 0.98 & 217 \\
        Shoulder Press & 0.99 & 1.00 & 1.00 & 223 \\
        \hline
        \textbf{Accuracy} & \multicolumn{2}{c|}{} & \textbf{0.99} & 791 \\
        \textbf{Macro Avg} & 0.99 & 0.99 & 0.99 & 791 \\
        \textbf{Weighted Avg} & 0.99 & 0.99 & 0.99 & 791 \\
        \hline
    \end{tabular}
\end{table}

\begin{table}[H]
    \centering
    \caption{Classification Report for BiLSTM Model}
    \label{tab:bilstmclassification-report_}
    \begin{tabular}{|l|c|c|c|c|}
        \hline
        \textbf{Class} & \textbf{Precision} & \textbf{Recall} & \textbf{F1-Score} & \textbf{Support} \\
        \hline
        Barbell Biceps Curl & 0.99 & 0.98 & 0.99 & 193 \\
        Push-up & 1.00 & 0.99 & 1.00 & 158 \\
        Squat & 0.98 & 0.99 & 0.98 & 217 \\
        Shoulder Press & 1.00 & 1.00 & 1.00 & 223 \\
        \hline
        \textbf{Accuracy} & \multicolumn{2}{c|}{} & \textbf{0.99} & 791 \\
        \textbf{Macro Avg} & 0.99 & 0.99 & 0.99 & 791 \\
        \textbf{Weighted Avg} & 0.99 & 0.99 & 0.99 & 791 \\
        \hline
    \end{tabular}
\end{table}

\begin{figure}[H]
    \centering
    \begin{subfigure}[b]{0.48\textwidth}
        \centering
        \includegraphics[width=\textwidth]{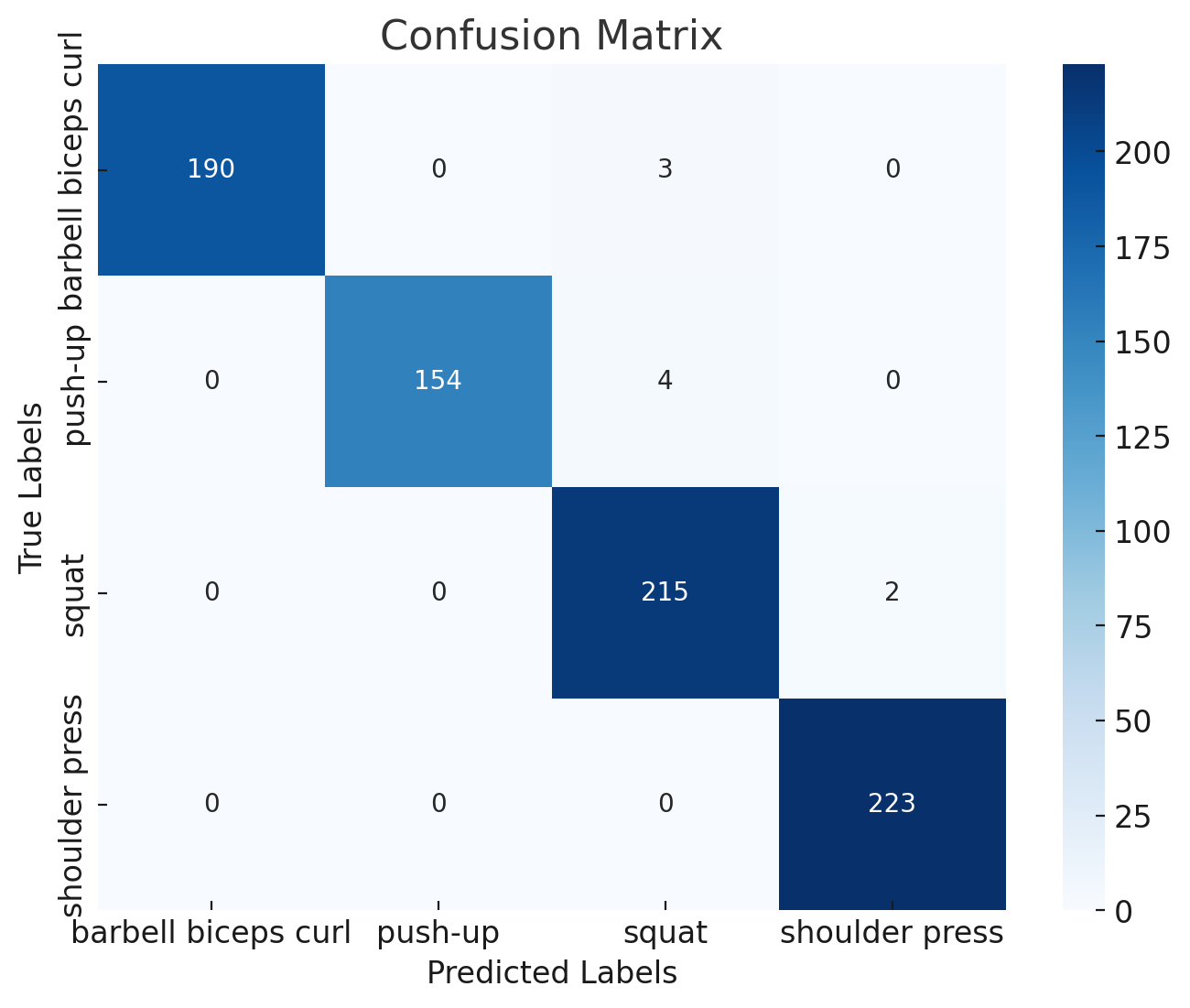}
        \caption{Confusion Matrix for LSTM Model}
        \label{fig:lstm-confusion-matrix}
    \end{subfigure}
    \hfill
    \begin{subfigure}[b]{0.48\textwidth}
        \centering
        \includegraphics[width=\textwidth]{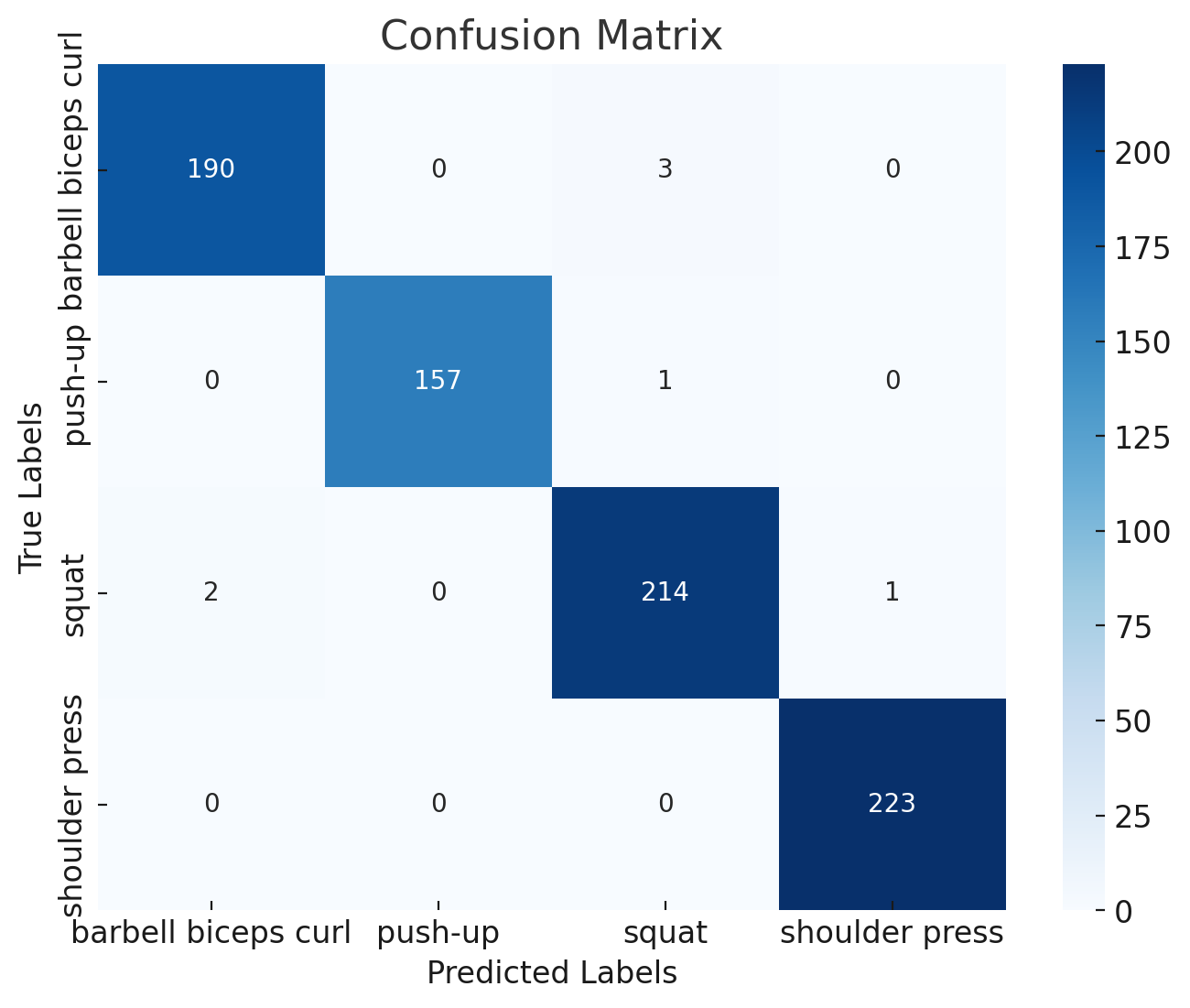}
        \caption{Confusion Matrix for BiLSTM Model}
        \label{fig:bilstm-confusion-matrix}
    \end{subfigure}
    \caption{Confusion Matrices for LSTM and BiLSTM Models}
    \label{fig:confusion-matrices}
\end{figure}

\begin{figure}[H]
    \centering
    \begin{subfigure}[b]{0.8\textwidth}
        \centering
        \includegraphics[width=\textwidth]{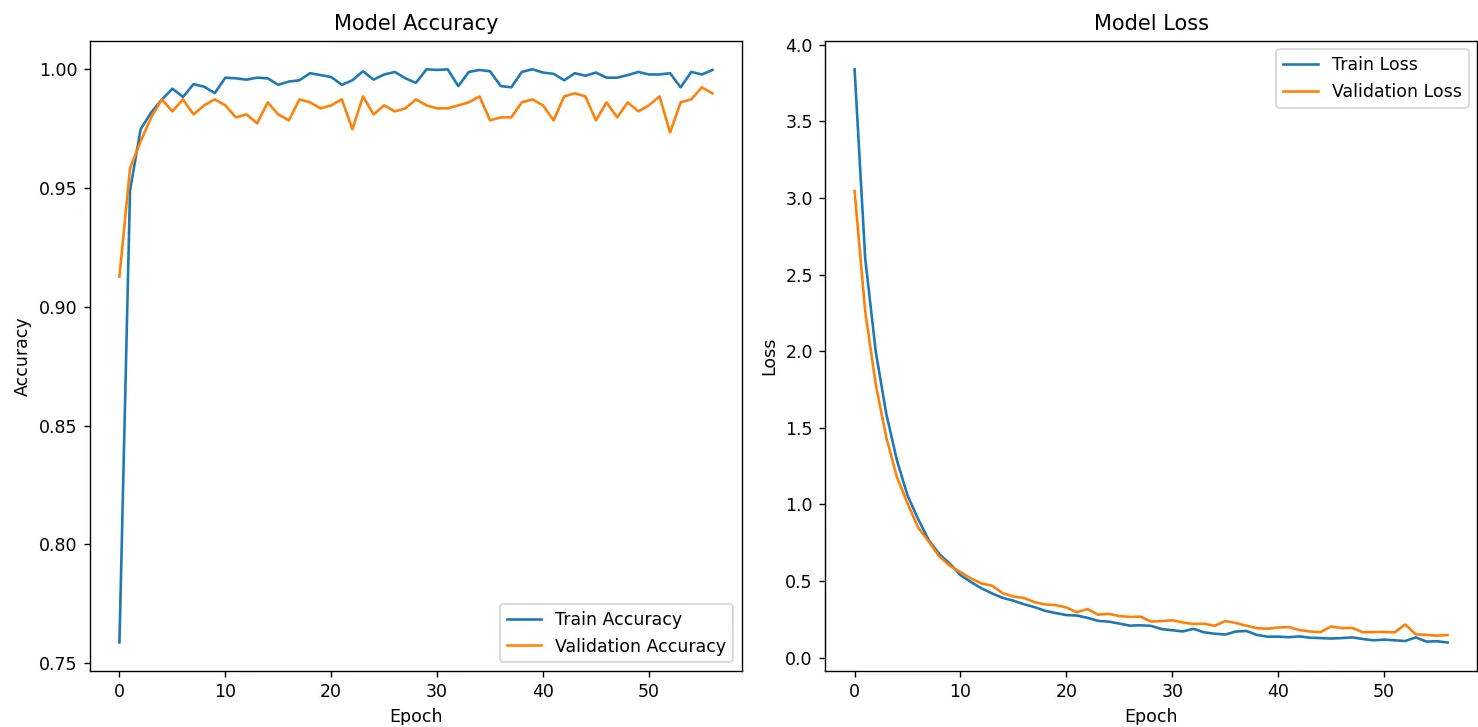}
        \caption{Learning Curve for LSTM Model}
        \label{fig:lstm-learning-curve}
    \end{subfigure}
    
    \vspace{1em} 

    \begin{subfigure}[b]{0.8\textwidth}
        \centering
        \includegraphics[width=\textwidth]{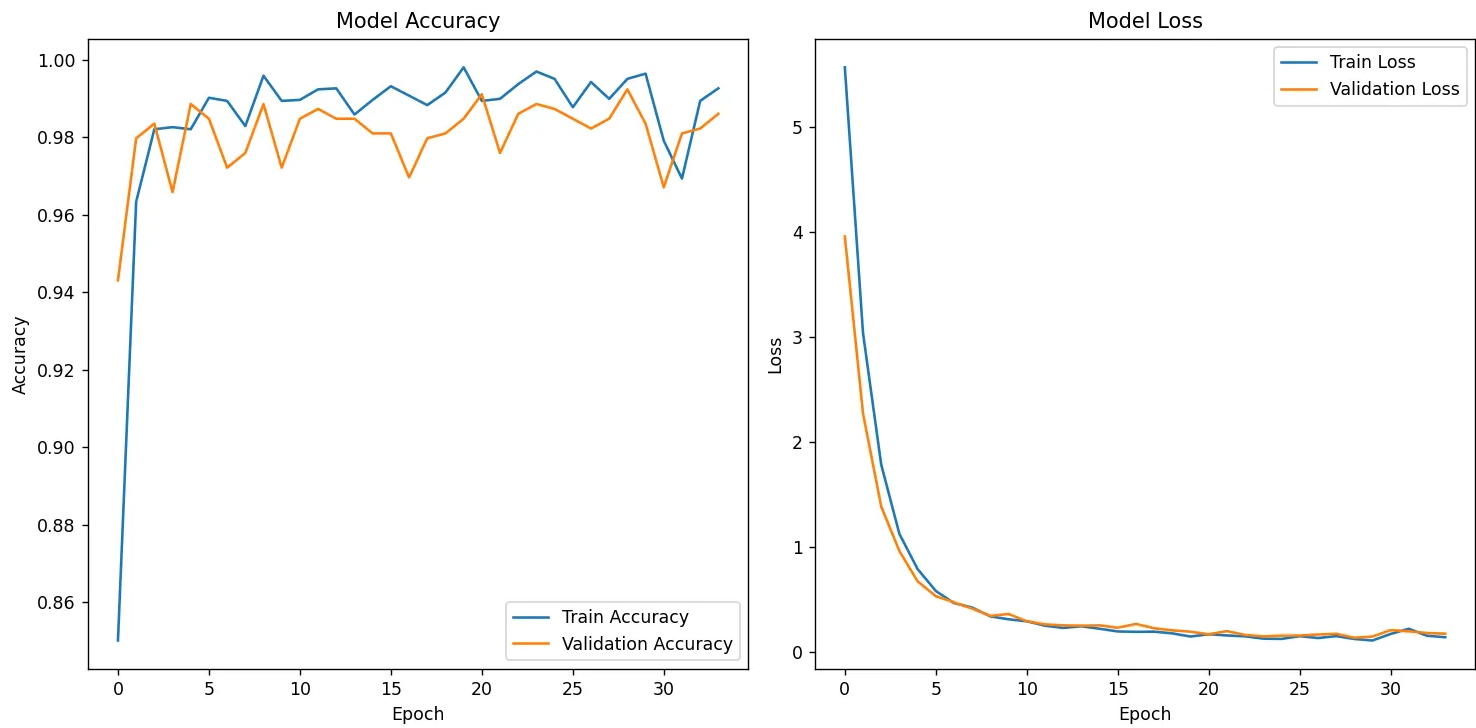}
        \caption{Learning Curve for BiLSTM Model}
        \label{fig:bilstm-learning-curve}
    \end{subfigure}

    \caption{Learning Curves for LSTM and BiLSTM Models}
    \label{fig:learning-curves--}
\end{figure}
\subsection{Evaluation on Additional Test Sets}

The additional test sets were used to assess the generalizability of the models. The first test set, “Final My Test Video,” consists of videos recorded under conditions recommended for the application (i.e., clear visibility of the head and body with a frontal or slightly angled view). The second test set, “Final Test Gym Video,” includes videos that do not strictly follow the recommended guidelines, such as those recorded in various environments like gyms, outdoors, or homes with different camera angles. The following table shows the accuracy and classification report for these test sets, along with the confusion matrix.\\

Here are the results for the dataset: Final My Test Video:

\begin{table}[H]
    \centering
    \caption{Classification Report for LSTM Model on Final My Test Video}
    \label{tab:lstm-test1-classification-report}
    \begin{tabular}{|l|c|c|c|c|}
        \hline
        \textbf{Class} & \textbf{Precision} & \textbf{Recall} & \textbf{F1-Score} & \textbf{Support} \\
        \hline
        Barbell Biceps Curl & 0.89 & 0.94 & 0.91 & 51 \\
        Push-up & 1.00 & 1.00 & 1.00 & 38 \\
        Squat & 0.90 & 0.95 & 0.92 & 58 \\
        Shoulder Press & 1.00 & 0.89 & 0.94 & 55 \\
        \hline
        \textbf{Accuracy} & \multicolumn{2}{c|}{} & 0.94 & 202 \\
        \textbf{Macro Avg} & 0.95 & 0.95 & 0.95 & 202 \\
        \textbf{Weighted Avg} & 0.94 & 0.94 & 0.94 & 202 \\
        \hline
    \end{tabular}
\end{table}

\begin{table}[H]
    \centering
    \caption{Classification Report for BiLSTM Model on Final My Test Video}
    \label{tab:bilstm-test1-classification-report__}
    \begin{tabular}{|l|c|c|c|c|}
        \hline
        \textbf{Class} & \textbf{Precision} & \textbf{Recall} & \textbf{F1-Score} & \textbf{Support} \\
        \hline
        Barbell Biceps Curl & 0.89 & 1.00 & 0.94 & 51 \\
        Push-up & 1.00 & 1.00 & 1.00 & 38 \\
        Squat & 0.93 & 0.93 & 0.93 & 58 \\
        Shoulder Press & 1.00 & 0.89 & 0.94 & 55 \\
        \hline
        \textbf{Accuracy} & \multicolumn{2}{c|}{} & 0.95 & 202 \\
        \textbf{Macro Avg} & 0.96 & 0.96 & 0.95 & 202 \\
        \textbf{Weighted Avg} & 0.95 & 0.95 & 0.95 & 202 \\
        \hline
    \end{tabular}
\end{table}

\begin{figure}[H]
    \centering
    \begin{subfigure}[b]{0.49\textwidth}
        \centering
        \includegraphics[width=\textwidth]{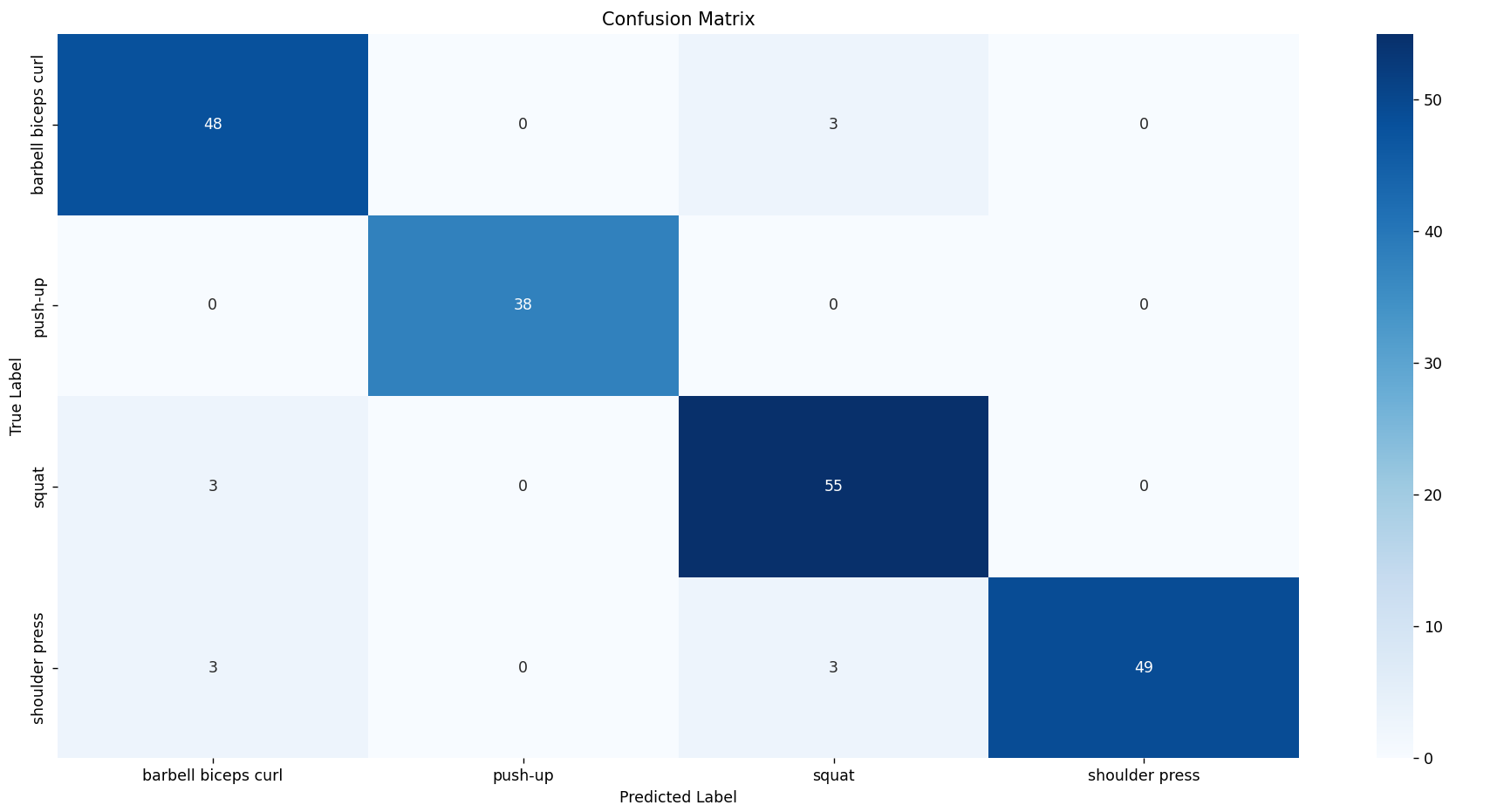}
        \caption{Confusion Matrix for LSTM Model}
        \label{fig:lstm-confusion-matrix_}
    \end{subfigure}
    \hfill
    \begin{subfigure}[b]{0.49\textwidth}
        \centering
        \includegraphics[width=\textwidth]{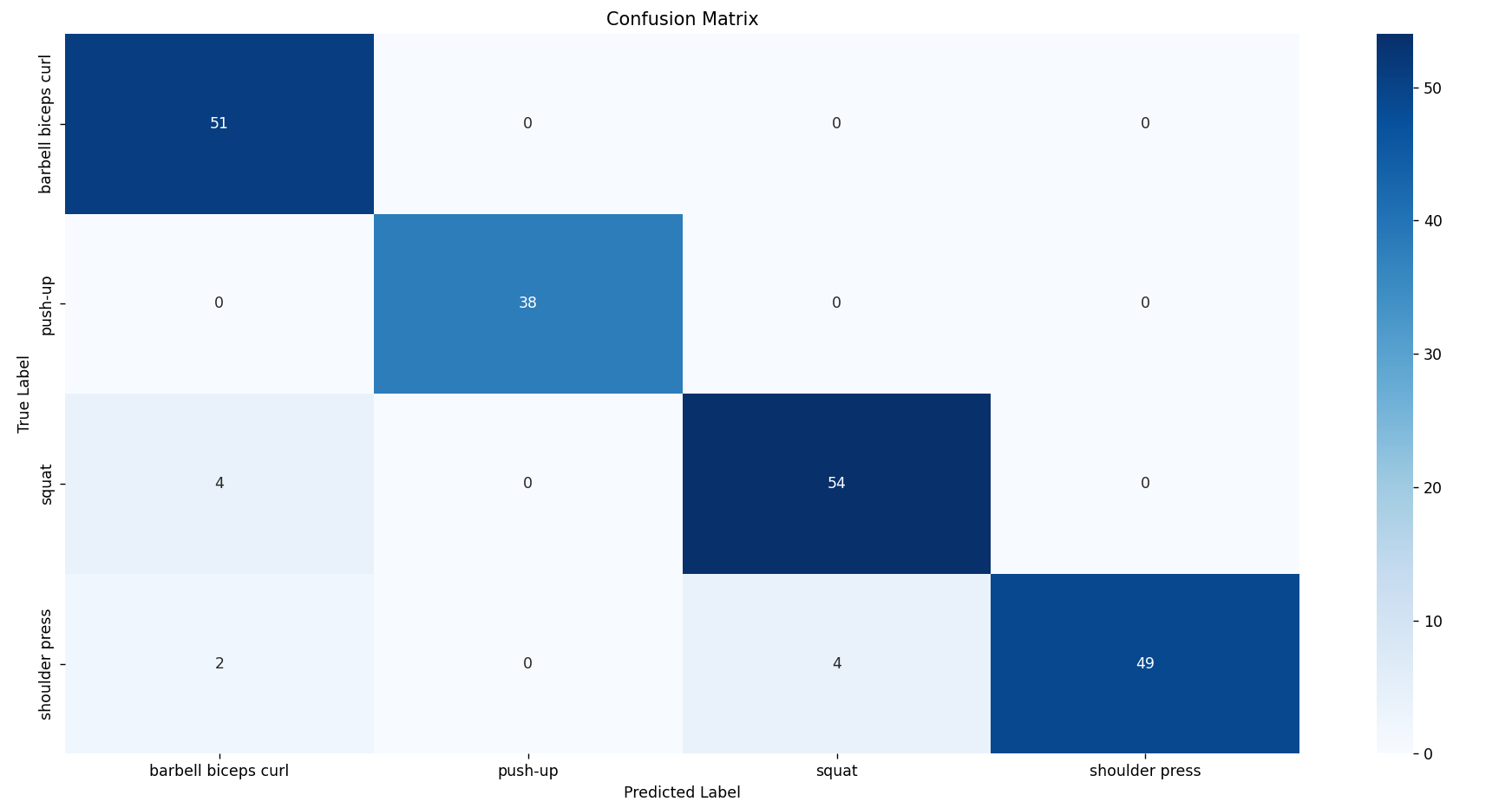}
        \caption{Confusion Matrix for BiLSTM Model}
        \label{fig:bilstm-confusion-matrix_}
    \end{subfigure}
    \caption{Confusion Matrices for LSTM and BiLSTM Models}
    \label{fig:confusion-matrices_}
\end{figure}

Here are the results for the dataset: Final Test Gym Video:

\begin{table}[H]
    \centering
    \caption{Classification Report for LSTM Model on Final Test Gym Video}
    \label{tab:lstm-test2-classification-report}
    \begin{tabular}{|l|c|c|c|c|}
        \hline
        \textbf{Class} & \textbf{Precision} & \textbf{Recall} & \textbf{F1-Score} & \textbf{Support} \\
        \hline
        Barbell Biceps Curl & 0.84 & 0.73 & 0.78 & 90 \\
        Push-up & 1.00 & 0.99 & 0.99 & 70 \\
        Squat & 0.63 & 0.84 & 0.72 & 58 \\
        Shoulder Press & 1.00 & 0.91 & 0.95 & 88 \\
        \hline
        \textbf{Accuracy} & \multicolumn{2}{c|}{} & 0.86 & 306 \\
        \textbf{Macro Avg} & 0.87 & 0.87 & 0.86 & 306 \\
        \textbf{Weighted Avg} & 0.88 & 0.86 & 0.87 & 306 \\
        \hline
    \end{tabular}
\end{table}

\begin{table}[H]
    \centering
    \caption{Classification Report for BiLSTM Model on Final Test Gym Video}
    \label{tab:bilstmtest2classification-report}
    \begin{tabular}{|l|c|c|c|c|}
        \hline
        \textbf{Class} & \textbf{Precision} & \textbf{Recall} & \textbf{F1-Score} & \textbf{Support} \\
        \hline
        Barbell Biceps Curl & 0.85 & 0.81 & 0.83 & 90 \\
        Push-up & 1.00 & 0.97 & 0.99 & 70 \\
        Squat & 0.67 & 0.83 & 0.74 & 58 \\
        Shoulder Press & 1.00 & 0.91 & 0.95 & 88 \\
        \hline
        \textbf{Accuracy} & \multicolumn{2}{c|}{} & 0.88 & 306 \\
        \textbf{Macro Avg} & 0.88 & 0.88 & 0.88 & 306 \\
        \textbf{Weighted Avg} & 0.89 & 0.88 & 0.88 & 306 \\
        \hline
    \end{tabular}
\end{table}

\begin{figure}[h]
    \centering
    \begin{subfigure}[b]{0.49\textwidth}
        \centering
        \includegraphics[width=\textwidth]{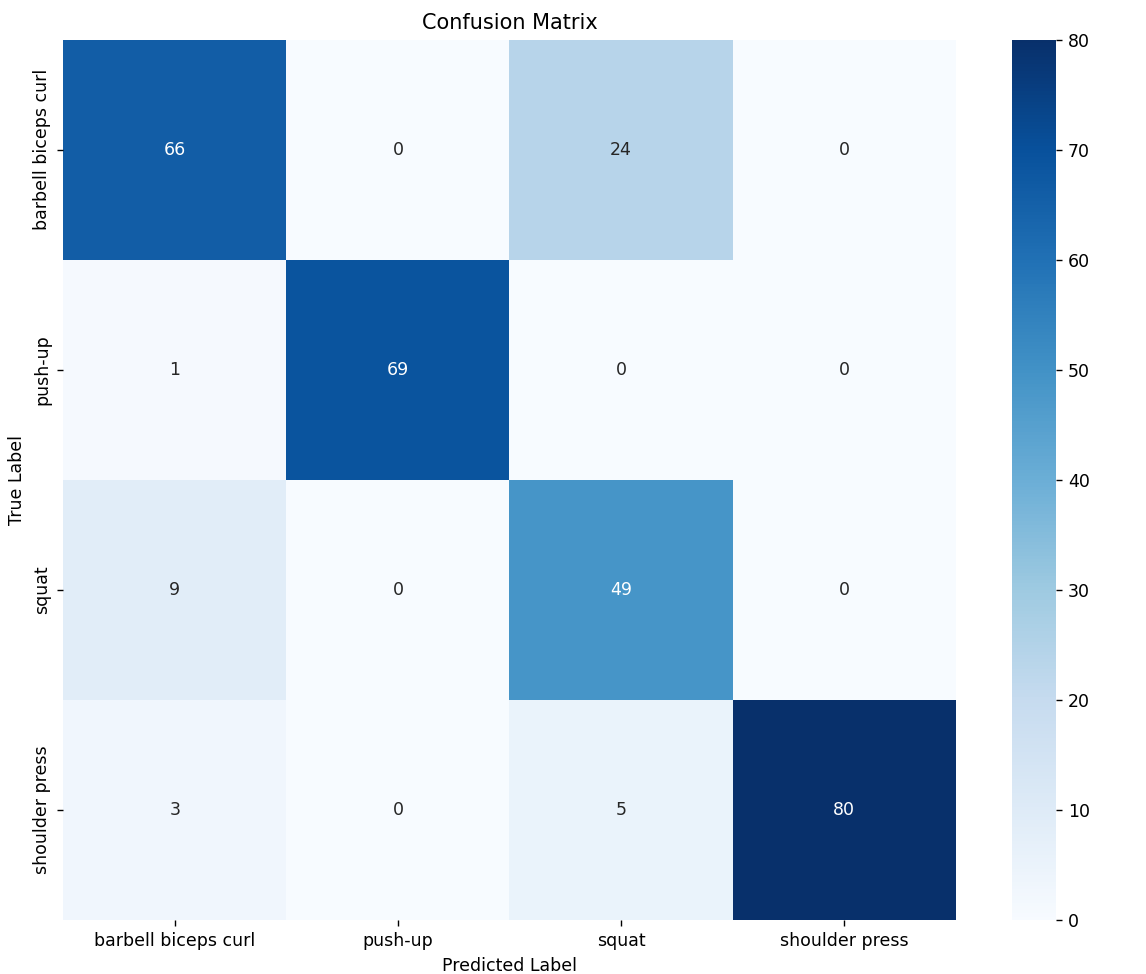}
        \caption{Confusion Matrix for LSTM Model}
        \label{fig:lstm-confusion-matrix__}
    \end{subfigure}
    \hfill
    \begin{subfigure}[b]{0.49\textwidth}
        \centering
        \includegraphics[width=\textwidth]{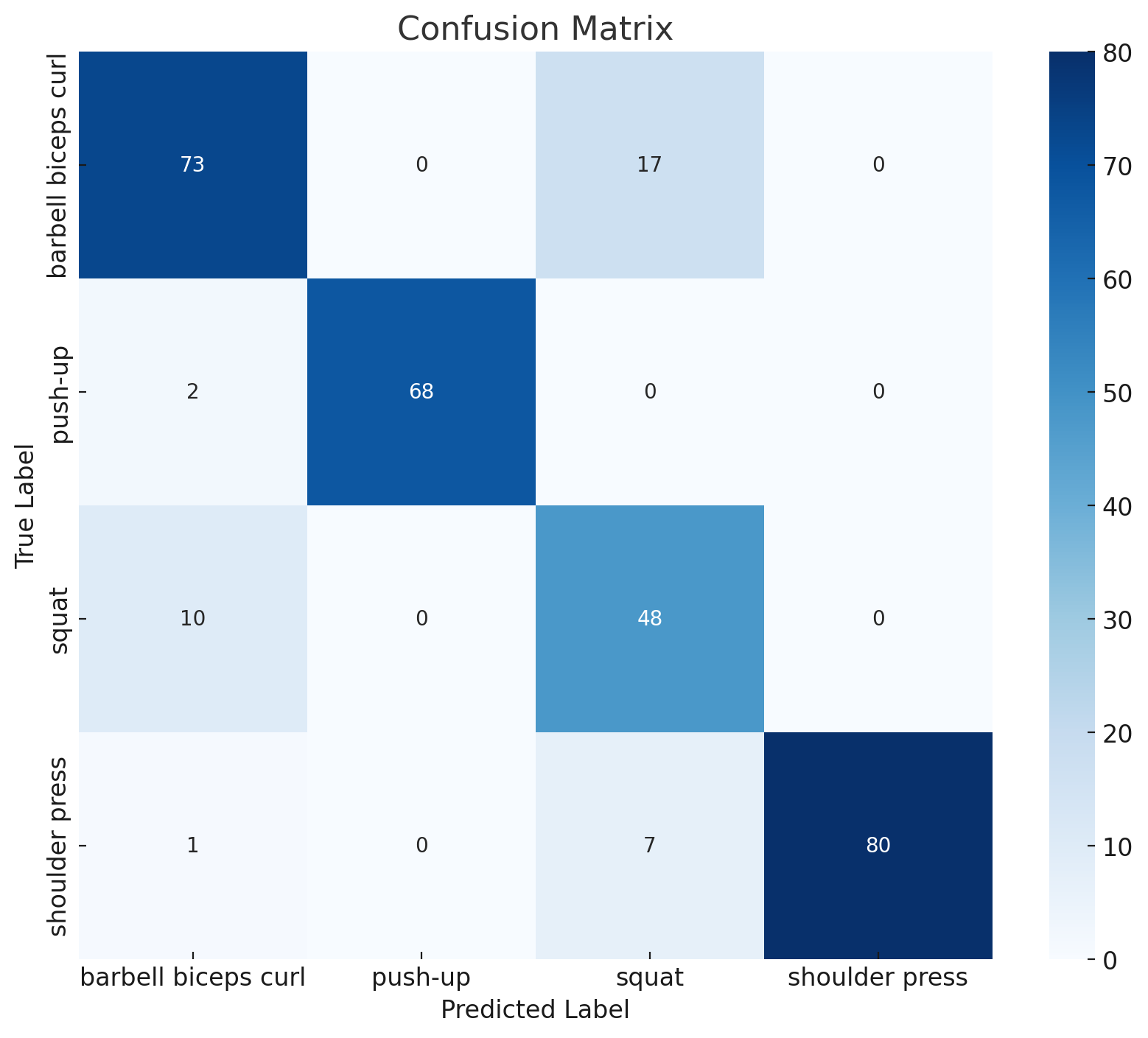}
        \caption{Confusion Matrix for BiLSTM Model}
        \label{fig:bilstm-confusion-matrix__}
    \end{subfigure}
    \caption{Confusion Matrices for LSTM and BiLSTM Models}
    \label{fig:confusion-matrices__}
\end{figure}

From this result, it can be seen that the two models are very close in terms of results with the BiLSTM having slightly superior performance and generalization capabilities on the additional test sets. From now on only this model will be used for further comparison.

\subsection{Model Comparison and Evaluation on Different Feature Types}

In addition to evaluating the LSTM and BiLSTM models on the test datasets, further experiments are conducted in order to examine the effect of using angles versus raw coordinates as input features. Specifically, a BiLSTM model was trained using the same architecture but just with the raw coordinates from the 33 landmarks detected by MediaPipe.
In addition, another model called "BiLSTM Invariant", which makes no use of raw coordinates but just uses invariant features like angle and normalized distances, is tested. 

\begin{table}[H]
\centering
\scriptsize 
\caption{Complete List of Features of BiLSTM Invariant}
\begin{tabular}{|l|l|}
\hline
\textbf{Feature Type} & \textbf{Landmarks} \\ \hline
\multirow{8}{*}{\textbf{Angles}} 
& LEFT\_SHOULDER, LEFT\_ELBOW, LEFT\_WRIST \\ 
& RIGHT\_SHOULDER, RIGHT\_ELBOW, RIGHT\_WRIST \\ 
& LEFT\_HIP, LEFT\_KNEE, LEFT\_ANKLE \\ 
& RIGHT\_HIP, RIGHT\_KNEE, RIGHT\_ANKLE \\ 
& LEFT\_SHOULDER, LEFT\_HIP, LEFT\_KNEE \\ 
& RIGHT\_SHOULDER, RIGHT\_HIP, RIGHT\_KNEE \\ 
& LEFT\_HIP, LEFT\_SHOULDER, LEFT\_ELBOW \\ 
& RIGHT\_HIP, RIGHT\_SHOULDER, RIGHT\_ELBOW \\ \hline

\multirow{11}{*}{\textbf{Normalized Distances}} 
& LEFT\_SHOULDER, RIGHT\_SHOULDER \\ 
& LEFT\_HIP, RIGHT\_HIP \\ 
& LEFT\_HIP, LEFT\_KNEE \\ 
& RIGHT\_HIP, RIGHT\_KNEE \\ 
& LEFT\_SHOULDER, LEFT\_HIP \\ 
& RIGHT\_SHOULDER, RIGHT\_HIP \\ 
& LEFT\_ELBOW, LEFT\_KNEE \\ 
& RIGHT\_ELBOW, RIGHT\_KNEE \\ 
& LEFT\_WRIST, LEFT\_SHOULDER \\ 
& RIGHT\_WRIST, RIGHT\_SHOULDER \\ 
& LEFT\_WRIST, LEFT\_HIP \\ 
& RIGHT\_WRIST, RIGHT\_HIP \\ \hline

\end{tabular}
\label{tab:features}
\end{table}

The results on the "Final Test Gym" and "Final My Test Video" datasets, presented in Tables 12 and 13, illustrate the impact of using raw coordinates instead of angle-based features. When angles are clearly visible, such as in a controlled environment, the angle-based model achieves significantly better results. However, when angles are not as visible (e.g., due to occlusion or varied camera angles), the coordinate-based model performs better. These findings suggest that a combined approach leveraging both features may provide optimal performance across diverse scenarios.

\begin{table}[H]
    \centering
    \caption{Classification Report for BiLSTM Model with Raw Coordinates on Final My Test Video}
    \label{tab:coord-test2}
    \begin{tabular}{|l|c|c|c|c|}
        \hline
        \textbf{Class} & \textbf{Precision} & \textbf{Recall} & \textbf{F1-Score} & \textbf{Support} \\
        \hline
        Barbell Biceps Curl & 0.93 & 0.98 & 0.96 & 55 \\
        Push-up & 0.64 & 1.00 & 0.78 & 38 \\
        Shoulder Press & 1.00 & 0.57 & 0.73 & 54 \\
        Squat & 0.64 & 0.63 & 0.63 & 59 \\
        \hline
        \textbf{Accuracy} & \multicolumn{2}{c|}{} & \textbf{0.78} & 206 \\
        \textbf{Macro Avg} & 0.80 & 0.80 & 0.78 & 206 \\
        \textbf{Weighted Avg} & 0.81 & 0.78 & 0.77 & 206 \\
        \hline
    \end{tabular}
\end{table}

\begin{table}[H]
    \centering
    \caption{Classification Report for BiLSTM Model with Raw Coordinates on Final Test Gym Video}
    \label{tab:coord-test1}
    \begin{tabular}{|l|c|c|c|c|}
        \hline
        \textbf{Class} & \textbf{Precision} & \textbf{Recall} & \textbf{F1-Score} & \textbf{Support} \\
        \hline
        Barbell Biceps Curl & 0.81 & 0.87 & 0.84 & 94 \\
        Push-up & 0.99 & 0.97 & 0.98 & 73 \\
        Shoulder Press & 1.00 & 0.93 & 0.96 & 87 \\
        Squat & 0.73 & 0.73 & 0.73 & 60 \\
        \hline
        \textbf{Accuracy} & \multicolumn{2}{c|}{} & \textbf{0.89} & 314 \\
        \textbf{Macro Avg} & 0.88 & 0.88 & 0.88 & 314 \\
        \textbf{Weighted Avg} & 0.89 & 0.89 & 0.89 & 314 \\
        \hline
    \end{tabular}
\end{table}

\vspace{1em}

The following tables show result for the BiLSTM Invariant:

\begin{table}[H]
    \centering
    \caption{Classification Report for BiLSTM Invariant Model on Final My Test Video}
    \label{tab:bilstm-test1classification-report}
    \begin{tabular}{|l|c|c|c|c|}
        \hline
        \textbf{Class} & \textbf{Precision} & \textbf{Recall} & \textbf{F1-Score} & \textbf{Support} \\
        \hline
        Barbell Biceps Curl & 0.80 & 0.96 & 0.88 & 55 \\
        Push-up & 0.94 & 0.87 & 0.90 & 38 \\
        Shoulder Press & 0.98 & 0.89 & 0.93 & 54 \\
        Squat & 0.80 & 0.76 & 0.78 & 59 \\
        \hline
        \textbf{Accuracy} & \multicolumn{2}{c|}{} & 0.87 & 206 \\
        \textbf{Macro Avg} & 0.88 & 0.87 & 0.87 & 206 \\
        \textbf{Weighted Avg} & 0.88 & 0.87 & 0.87 & 206 \\
        \hline
    \end{tabular}
\end{table}

\begin{table}[H]
    \centering
    \caption{Classification Report for BiLSTM Invariant Model on Final Test Gym Video}
    \label{tab:bilstm-test2-classification-report}
    \begin{tabular}{|l|c|c|c|c|}
        \hline
        \textbf{Class} & \textbf{Precision} & \textbf{Recall} & \textbf{F1-Score} & \textbf{Support} \\
        \hline
        Barbell Biceps Curl & 0.75 & 0.79 & 0.77 & 94 \\
        Push-up & 0.95 & 0.79 & 0.87 & 73 \\
        Shoulder Press & 1.00 & 0.85 & 0.92 & 87 \\
        Squat & 0.62 & 0.83 & 0.71 & 60 \\
        \hline
        \textbf{Accuracy} & \multicolumn{2}{c|}{} & 0.82 & 314 \\
        \textbf{Macro Avg} & 0.83 & 0.82 & 0.82 & 314 \\
        \textbf{Weighted Avg} & 0.84 & 0.82 & 0.82 & 314 \\
        \hline
    \end{tabular}
\end{table}

\subsection{Comparison with previous approaches  }
One challenge faced in this study, and in the broader field of exercise classification, is the absence of a standardized benchmark dataset. Without a common dataset used across studies, it becomes difficult to directly compare the performance of different models. Existing approaches often rely on proprietary or specific datasets, each with unique characteristics that may not consistently reflect real-world conditions. The lack of a widely adopted benchmark hinders the ability to measure progress across studies effectively. This is one reason why the model was integrated into a real-time fitness application, allowing for practical evaluation in real-world settings where users engage with the system directly. Testing the model in the app offers valuable insights into how it performs under various conditions, supplementing the gaps left by the lack of standardized datasets. Future research should consider the development of a standardized dataset for exercise classification, which would enable more reliable comparisons and encourage further advancements in this domain.
Keeping in mind the problem of a benchmark dataset, this paper compared the proposed model with the previous approaches by implementing their model architecture and training and testing on the dataset used for evaluating the proposed model. In particular, [4] and [6] were implemented, while [5] was not directly implemented since previous results already demonstrated the superiority of BiLSTM over LSTM, as well as the advantages of combining angle and coordinate features over using raw coordinates alone. Below are reported the results of the model implemented and discussed some choices regarding their implementation. In all implementations, hyperparameter tuning has been used, specifically tuning the learning rate, batch size, and number of epochs, as in the proposed model.
For [4], the exact architecture described in the paper was used. While the precise angle features they utilized were not explicitly detailed, it can be inferred from the text that they employed similar features used in the proposed model, so these features were used. Additionally, it was unclear whether they used a sliding window or a non-overlapping window for generating predictions on 30-frame sequences. However, this distinction is less critical since the training was conducted at the individual frame level. Both methods were implemented and produced nearly identical results, with the non-overlapping approach ultimately chosen to maintain consistency for comparison purposes.\\
\\
Here are the results of the CNN+ soft voting method:

\begin{table}[H]
    \centering
    \caption{Classification Report for Frame-Level and Sequence-Level Performance}
    \label{tab:classificationreport}
    \begin{tabular}{|l|c|c|c|c|}
        \hline
        \textbf{Class} & \textbf{Precision} & \textbf{Recall} & \textbf{F1-Score} & \textbf{Support} \\
        \hline
        \multicolumn{5}{|c|}{\textbf{Frame-Level}} \\
        \hline
        Barbell Biceps Curl & 0.90 & 0.95 & 0.93 & 7251 \\
        Push-up & 1.00 & 0.99 & 1.00 & 5165 \\
        Squat & 0.94 & 0.86 & 0.89 & 6197 \\
        Shoulder Press & 0.97 & 0.99 & 0.98 & 7833 \\
        \hline
        \textbf{Accuracy} & \multicolumn{2}{c|}{} & 0.95 & 26446 \\
        \textbf{Macro Avg} & 0.95 & 0.95 & 0.95 & 26446 \\
        \textbf{Weighted Avg} & 0.95 & 0.95 & 0.95 & 26446 \\
        \hline
        \multicolumn{5}{|c|}{\textbf{Sequence-Level}} \\
        \hline
        Barbell Biceps Curl & 0.93 & 0.98 & 0.65 & 232 \\
        Push-up & 1.00 & 1.00 & 1.00 & 163 \\
        Squat & 0.97 & 0.90 & 0.94 & 197 \\
        Shoulder Press & 1.00 & 1.00 & 0 & 251 \\
        \hline
        \textbf{Accuracy} & \multicolumn{2}{c|}{} & 0.97 & 843 \\
        \textbf{Macro Avg} & 0.97 & 0.97 & 0.97 & 843 \\
        \textbf{Weighted Avg} & 0.97 & 0.97 & 0.97 & 843 \\
        \hline
    \end{tabular}
\end{table}

\begin{table}[H]
    \centering
    \caption{Sequence-Level Classification Report for Final My Test Video and Final Test Gym}
    \label{tab:sequence-level-comparison_}
    \resizebox{\textwidth}{!}{%
    \begin{tabular}{|l|c|c|c|c|c|c|c|c|}
        \hline
        \textbf{Class} & \multicolumn{4}{c|}{\textbf{Final My Test Video}} & \multicolumn{4}{c|}{\textbf{Final Test Gym}} \\
        \hline
        & \textbf{Precision} & \textbf{Recall} & \textbf{F1-Score} & \textbf{Support} 
        & \textbf{Precision} & \textbf{Recall} & \textbf{F1-Score} & \textbf{Support} \\
        \hline
        Barbell Biceps Curl & 0.82 & 1.00 & 0.90 & 51 
                           & 0.72 & 0.99 & 0.84 & 90 \\
        Push-Up             & 0.73 & 1.00 & 0.84 & 38 
                           & 1.00 & 0.94 & 0.97 & 76 \\
        Squat               & 1.00 & 0.67 & 0.80 & 58 
                           & 0.97 & 0.50 & 0.66 & 58 \\
        Shoulder Press      & 1.00 & 0.89 & 0.94 & 55 
                           & 1.00 & 0.99 & 0.99 & 88 \\
        \hline
        \textbf{Accuracy}   & \multicolumn{2}{c|}{} & 0.88 & 202 
                           & \multicolumn{2}{c|}{} & 0.89 & 306 \\
        \textbf{Macro Avg}  & 0.89 & 0.89 & 0.87 & 202 
                           & 0.92 & 0.86 & 0.86 & 306 \\
        \textbf{Weighted Avg} & 0.90 & 0.88 & 0.87 & 202 
                              & 0.91 & 0.89 & 0.88 & 306 \\
        \hline
    \end{tabular}%
    }
\end{table}

For the paper [6], the architecture remained exactly the same; the only differences were the use of MediaPipe instead of OpenPose to ensure a fair comparison and the substitution of a point landmark. Specifically, the landmark located between the shoulders in OpenPose was replaced by the left mouth landmark in MediaPipe. In this case, a sliding window of 1 frame, as explicitly stated in the paper, was used.\\
\\
Here are the results of the DNN+ majority voting:
\begin{table}[H]
    \centering
    \caption{Classification Report for Frame-Level and Sequence-Level Performance}
    \label{tab:classification-reports}
    \begin{tabular}{|l|c|c|c|c|}
        \hline
        \textbf{Class} & \textbf{Precision} & \textbf{Recall} & \textbf{F1-Score} & \textbf{Support} \\
        \hline
        \multicolumn{5}{|c|}{\textbf{Frame-Level}} \\
        \hline
        Barbell Biceps Curl & 0.91 & 0.84 & 0.87 & 6817 \\
        Push-up & 1.00 & 1.00 & 1.00 & 5166 \\
        Squat & 0.99 & 0.99 & 0.99 & 6904 \\
        Shoulder Press & 0.83 & 0.91 & 0.97 & 5965 \\
        \hline
        \textbf{Accuracy} & \multicolumn{2}{c|}{} & 0.93 & 25202 \\
        \textbf{Macro Avg} & 0.93 & 0.93 & 0.93 & 25202 \\
        \textbf{Weighted Avg} & 0.93 & 0.93 & 0.93 & 25202 \\
        \hline
        \multicolumn{5}{|c|}{\textbf{Sequence-Level}} \\
        \hline
        Barbell Biceps Curl & 0.92 & 0.84 & 0.87 & 6628 \\
        Push-up & 1.00 & 1.00 & 1.00 & 5336 \\
        Squat & 0.99 & 0.99 & 0.99 & 6715 \\
        Shoulder Press & 0.83 & 0.92 & 0.87 & 5803 \\
        \hline
        \textbf{Accuracy} & \multicolumn{2}{c|}{} & 0.93 & 24482 \\
        \textbf{Macro Avg} & 0.93 & 0.93 & 0.93 & 24482 \\
        \textbf{Weighted Avg} & 0.93 & 0.93 & 0.93 & 24482 \\
        \hline
    \end{tabular}
\end{table}

\begin{table}[H]
    \centering
    \caption{Sequence-Level Classification Report for Final My Test Video and Final Test Gym}
    \label{tab:sequence-level-comparison__}
    \resizebox{\textwidth}{!}{%
    \begin{tabular}{|l|c|c|c|c|c|c|c|c|}
        \hline
        \textbf{Class} & \multicolumn{4}{c|}{\textbf{Final My Test Video}} & \multicolumn{4}{c|}{\textbf{Final Test Gym}} \\
        \hline
        & \textbf{Precision} & \textbf{Recall} & \textbf{F1-Score} & \textbf{Support} 
        & \textbf{Precision} & \textbf{Recall} & \textbf{F1-Score} & \textbf{Support} \\
        \hline
        Barbell Biceps Curl & 0.55 & 0.98 & 0.71 & 1578 
                           & 0.72 & 0.89 & 0.79 & 2767 \\
        Push-up             & 0.88 & 0.96 & 0.91 & 1147 
                           & 0.98 & 0.91 & 0.95 & 2156 \\
        Shoulder Press      & 0.94 & 0.63 & 0.75 & 1596 
                           & 0.95 & 0.98 & 0.97 & 2538 \\
        Squat               & 0.95 & 0.53 & 0.68 & 1773 
                           & 0.75 & 0.50 & 0.60 & 1799 \\
        \hline
        \textbf{Accuracy}   & & & 0.75 & 6094
                           & & & 0.84 & 9300\\
        \textbf{Macro Avg}  & 0.83 & 0.77 & 0.76 & 6094 
                           & 0.85 & 0.82 & 0.83 & 9300 \\
        \textbf{Weighted Avg} & 0.83 & 0.75 & 0.75 & 6094 
                              & 0.85 & 0.84 & 0.84 & 9300 \\
        \hline
    \end{tabular}%
    }
\end{table}

\subsection{Summary of Results}

The following table reports the accuracy of the 3 test sets for the various models analyzed.

\begin{table}[H]
    \centering
    \caption{Accuracy of Models on Test Set, My Test Video, and Final Test Gym}
    \label{tab:model-accuracy-comparison}
    \renewcommand{\arraystretch}{1.5} 
    \resizebox{\textwidth}{!}{%
    \begin{tabular}{|l|c|c|c|}
        \hline
        \textbf{Model} & \textbf{Test Set Accuracy} & \textbf{My Test Video Accuracy} & \textbf{Final Test Gym Accuracy} \\
        \hline
        \textbf{BiLSTM with Mixed Features} & 0.9924 & 0.9505 & 0.8791 \\
        LSTM with Mixed Features & 0.9924 & 0.9406 & 0.8627 \\
        BiLSTM with Raw Coordinate & NA & 0.7800 & 0.8900 \\
        BiLSTM Invariant & 0.9823 & 0.8600 & 0.8100 \\
        DNN with Majority Voting (on Sequence) & 0.9318 & 0.7519 & 0.8446 \\
        CNN with Soft Voting (on Sequence) & 0.9715 & 0.8762 & 0.8856 \\
        \hline
    \end{tabular}%
    }
\end{table}

\subsection{Considerations}

Both LSTM and BiLSTM perform very well on the main dataset, achieving 99\% accuracy, which suggests that the models are properly tuned and effective for the conditions of the training data. Their ability to generalize to more diverse environments, as seen in the additional test sets, maintains strong performance while decreasing the data with more occlusions and difficult angles. In general, the BiLSTM model performs better than LSTM in handling diversified test datasets (the difference is small). The BiLSTM has the advantage of being able to capture temporal dependencies in exercise sequences and hence build a more accurate representation of movements. From the extensive evaluation of both features and other architecture, it is concluded that the best set of features considers both raw coordinates and angle. Crucial was also the ability of the model to leverage sequential data with respect to models that learn from a single frame.

Beyond technical evaluation, a subjective evaluation of how the classification is in real time while using the application can be considered though it would require more review from multiple users. A preliminary review shows that the model works well overall, but there is a tendency for the first repetition of an exercise not to be counted when switching between exercises. This is primarily because the model needs to "observe" the first repetition in its entirety to accurately recognize which exercise is being performed. Future improvements might involve optimizing the model to identify exercises more quickly, potentially by reducing the number of frames required for prediction, thereby shortening the time before the exercise is recognized.

In conclusion, the developed models achieve high performance on the main dataset set and maintain good performance also on other diverse test sets. The usage in the app is smooth but reducing prediction time might be considered. Finally, expanding the dataset to include more diverse exercise contexts seems to be the critical step toward improving generalization.

\section{Overview of the Web App}

The following section provides a general overview of the Fitness AI web application, showing the main functionalities and how they are integrated into the overall structure of the app.

The project is designed as a web application built with Streamlit [16], aimed at providing users with fitness tools such as real-time exercise classification, repetition counting, and a chatbot for fitness guidance.

The application interface has a main navigation sidebar that allows users to navigate between four pages with different functionalities:

\begin{enumerate}
    \item \textbf{Video Analysis}: This feature enables users to upload videos of their exercises, select the type of exercise from a list, and count the repetitions of that exercise. The video analysis process involves pose estimation using MediaPipe to extract landmarks, which are then analyzed to detect specific angle movements corresponding to each exercise type and increase the counter based on that.
    
    \item \textbf{Webcam Mode}: In this mode, users can perform exercises in front of their webcam, and the application provides real-time repetition counting. The webcam mode is optimized for exercises that are performed directly in front of the camera, and it utilizes similar pose estimation and analysis techniques as the video analysis mode.
    
    \item \textbf{Auto Classify Mode}: This mode is designed for users who prefer to switch between different exercises during their workout without having to manually select each exercise type. The application uses a BiLSTM model to classify exercises in real time and automatically applies the appropriate repetition counting logic based on the identified exercise.
    
    \item \textbf{Chatbot}: The chatbot in the Fitness AI web application acts as a fitness coach, designed to assist users with their fitness-related questions. It’s configured with a specific role to behave as an expert fitness trainer. The chatbot utilizes conversational memory to maintain context and provide more personalized interactions. Additionally, a warning is displayed within the app, indicating that the chatbot may occasionally make errors, and its advice should be verified for important decisions.
\end{enumerate}

A demo of the Fitness AI web application is available in the associated GitHub repository [17].

\begin{figure}[H]
\centering
\includegraphics[ width=1\textwidth]{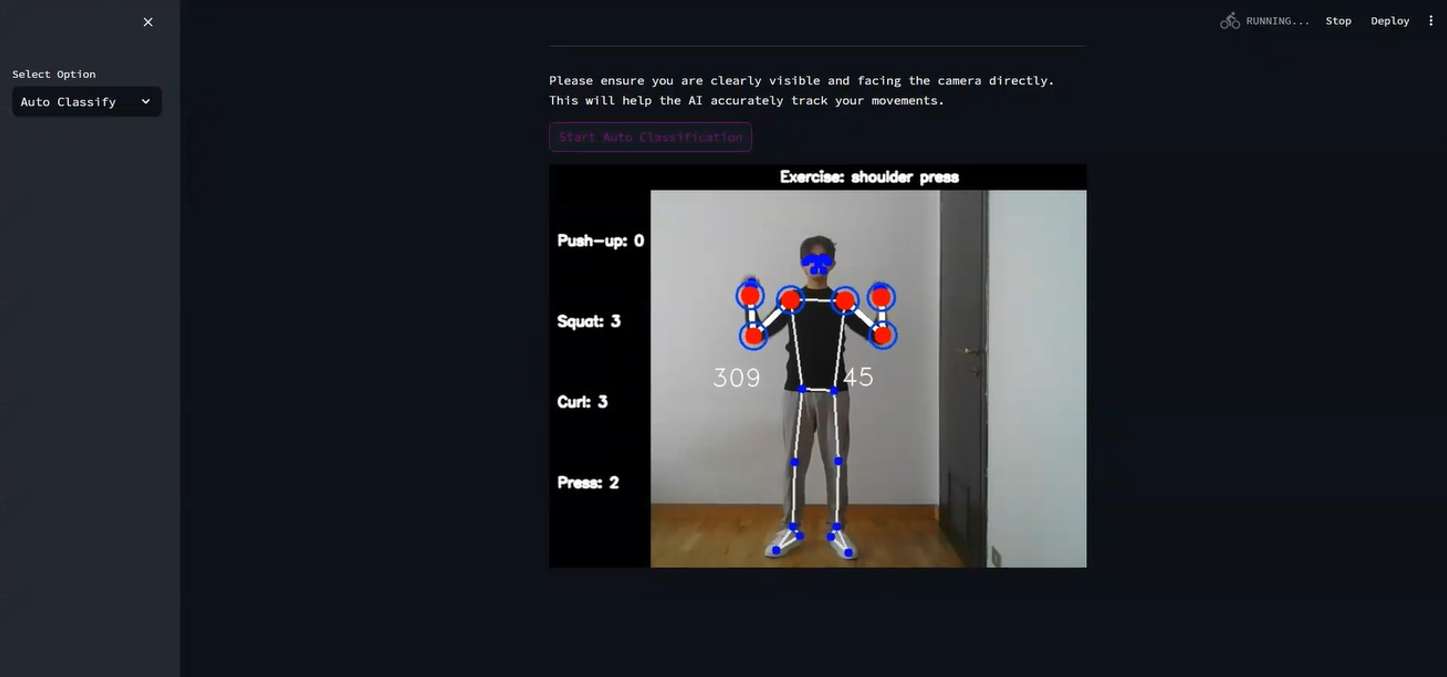} 
\caption{Auto Classify Mode page}
\label{fig:autoclassify}
\end{figure}

\subsection{Exercise Recognition in Practice}

The Auto Classify Mode serves as the primary showcase for the exercise recognition model. In this mode, the BiLSTM model processes a 30-frame window of pose data every second, classifying the exercise in real-time. This allows users to switch between exercises during their workout without manual input, demonstrating the model's adaptability and accuracy in a practical setting.

\subsection{Repetition Counting Implementation}

The repetition counting feature utilizes angle-based logic tailored to each exercise type. It tracks specific body landmarks and calculates angles between joints to determine the exercise stage (e.g., "up" or "down" position). The system counts a repetition when it detects a complete cycle of movement based on predefined angle thresholds. While effective, this approach could benefit from expert review to refine the angle ranges for optimal accuracy.

\subsection{Chatbot Implementation}

The chatbot feature uses OpenAI's GPT-3.5-turbo model [18], configured to act as a fitness expert. It provides users with an interactive way to get fitness-related information within the app. The chatbot maintains conversational context across interactions and includes a warning about potential inaccuracies.

While the additional features, including the chatbot and repetition counting logic, complement the core exercise classification model and enhance the user experience, their implementation details are not extensively discussed here as they are not the primary focus of this paper.

\section{Conclusion}

This paper introduced a novel approach for exercise classification and repetition counting using a BiLSTM model combined with pose estimation techniques. The proposed method addressed key limitations of existing systems, such as sensitivity to user positioning, camera angles, and variability in individual body types. By leveraging invariant features, such as joint angles, and utilizing the sequential nature of video data, the model demonstrated robustness in a variety of real-world conditions.

The model was trained on a dataset that combined synthetic and real-world data, achieving a Test accuracy of 99\% and a strong performance in other real-world test sets. The use of BiLSTM architecture enabled the model to capture the temporal context of exercises more effectively than previous methods, making it suitable for distinguishing between exercises that share similar initial postures but differ over time.

While the results are promising, the paper also identified certain limitations, particularly in the generalizability of the model to more diverse environments, such as gyms or outdoor settings with varied angulations and perspectives. This suggests that future efforts should focus on expanding the dataset with more diverse samples and further refining the model to handle these conditions more effectively. Additionally, the model was implemented inside a web application to be tested in a practical context.



\end{document}